\documentclass{article} 
\usepackage{iclr2027_conference,times}

\usepackage{amsmath,amsfonts,bm}

\def\eqref#1{equation~\ref{#1}}

\def\1{\bm{1}}

\DeclareMathAlphabet{\mathsfit}{\encodingdefault}{\sfdefault}{m}{sl}
\SetMathAlphabet{\mathsfit}{bold}{\encodingdefault}{\sfdefault}{bx}{n}

\usepackage{hyperref}
\usepackage{url}

\usepackage{algorithm}
\usepackage{multicol}
\usepackage{multirow}
\usepackage[table]{xcolor}
\usepackage{amssymb}
\usepackage{caption}
\usepackage{amsmath}
\usepackage{graphicx}
\usepackage{booktabs}
\usepackage{fontawesome5}
\usepackage{tabularx}
\usepackage{algpseudocode}
\usepackage{wrapfig}
\usepackage{marvosym}

\title{From Glance to Scrutiny: Progressive Distortion Reasoning for Fine-Grained Image Quality Assessment}

\author{%
Aoting Zhang\textsuperscript{1,4} \quad
Mingze Gao\textsuperscript{2} \quad
Dongbao Yang\textsuperscript{3} \quad
Longyi Chen\textsuperscript{2} \\[2pt]
\textbf{Daoxin Zhang}\textsuperscript{\textbf{2}} \quad
\textbf{Yi Wu}\textsuperscript{2} \quad
\textbf{Yao Hu}\textsuperscript{2} \quad
\textbf{Yu Zhou}\textsuperscript{3} \\[2pt]
\textsuperscript{1}IIE, Chinese Academy of Sciences \quad 
\textsuperscript{2}Xiaohongshu Inc. \\
\textsuperscript{3}Nankai University \quad
\textsuperscript{4}University of Chinese Academy of Sciences\\[2pt]
\texttt{zhangaoting@iie.ac.cn} \quad
\texttt{yangdongbao@nankai.edu.cn}
}

\iclrfinalcopy 
\begin{document}

\maketitle

\begin{abstract}
Multi-modal large language models (MLLMs) have demonstrated significant potential in image quality assessment (IQA) by bridging visual perception with descriptive evaluations. However, existing approaches mainly focus on holistic quality prediction, often functioning as black boxes that provide limited insight into where distortions occur and how they affect perceived quality, hindering fine-grained analysis of localized and heterogeneous degradations.
We propose GS-IQA, a framework that reformulates IQA as a progressive \textit{Where--What--How} diagnosis, emulating the human perceptual process from an initial glance to closer scrutiny. Since a severity judgment is meaningful only for a correctly localized and recognized region, we realize this progression through a two-stage reinforcement learning paradigm that respects such dependencies: the glance stage uses a perception-gated reward to establish where degradations lie and what they are, activating severity feedback only once both are correct, while the scrutiny stage introduces online reward-conditioned degradation generation to synthesize hard examples targeted at the model's perceptual bottlenecks, sharpening its discrimination of subtle severity variations. To enable systematic evaluation, we construct Diag-Bench, a region-level IQA benchmark of about 25K curated samples spanning 12 distortion types and five ordinal severity levels. Extensive experiments show that GS-IQA consistently surpasses state-of-the-art methods in distortion localization, recognition, and severity estimation, and that its diagnostic representations transfer effectively to conventional global quality prediction across diverse external benchmarks. Code and data will be released.
\end{abstract}

\section{Introduction}
\label{sec:intro}
Image Quality Assessment (IQA) aims to evaluate visual quality consistently with human perception. Depending on the availability of a pristine reference, existing methods are broadly categorized into full-reference (FR) and no-reference (NR) approaches. FR-IQA methods compare distorted images with their references using structural metrics or learned perceptual features~\cite{wang2004image, zhang2018unreasonable}, whereas NR-IQA methods estimate quality without references, evolving from natural scene statistics~\cite{mittal2012making, mittal2012no} to deep quality-aware models~\cite{su2020blindly, ding2020image, ke2021musiq, ding2021locally}. Despite strong numerical performance, their reliance on holistic score regression often renders the assessment process opaque, offering limited evidence about the distortions underlying a predicted quality score.

\begin{figure*}[t]
    \centering
    \includegraphics[width=0.93\linewidth]{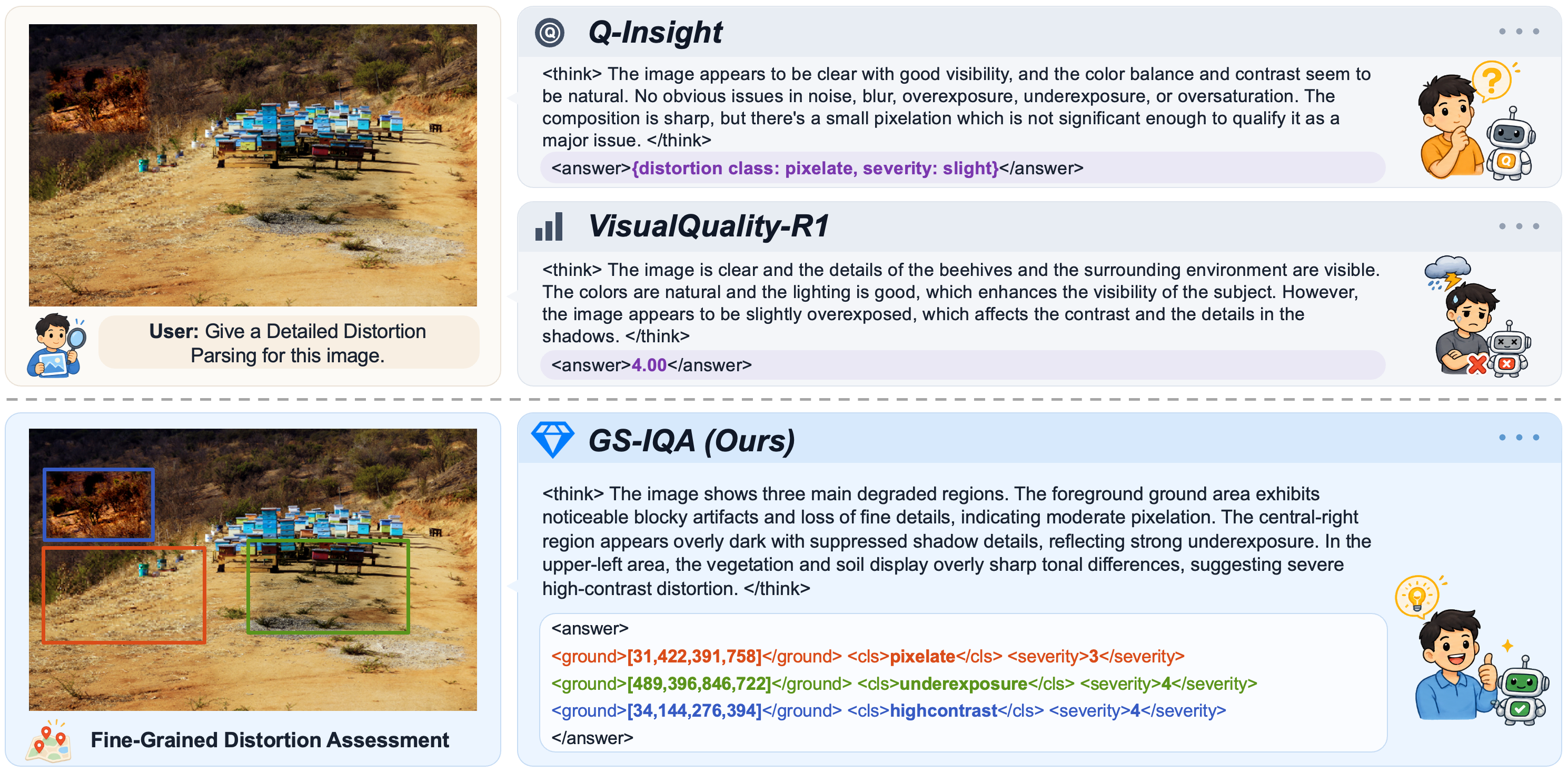}
    \caption{Qualitative comparison of GS-IQA with existing MLLM-based IQA methods. While Q-Insight and VisualQuality-R1 condense spatially heterogeneous degradations into a single distortion description or holistic quality score, GS-IQA employs progressive distortion reasoning to localize individual degraded regions and associate each with its distortion type and ordinal severity, enabling fine-grained and spatially grounded quality assessment.}
    \label{fig:illustration_iqa}
\vspace{-5mm}
\end{figure*}

The emergence of multimodal large language models (MLLMs)~\cite{liu2023visual, ye2024mplug} has extended IQA beyond scalar prediction toward language-based perceptual understanding. Existing approaches generally include score-based methods~\cite{wu2023q, you2025teaching}, which perform ordinal regression or preference ranking, and description-based methods~\cite{wu2024q, you2024depicting}, which explain perceived artifacts in natural language. Despite improved interpretability through instruction tuning~\cite{chen2024seagull} and large-scale quality-description corpora, most assessments remain image-level summaries, failing to explicitly associate distortion location, category, and severity. This limitation is particularly evident for spatially heterogeneous degradations.
As illustrated in Figure~\ref{fig:illustration_iqa}, Q-Insight~\cite{li2025qinsight} identifies only a dominant degradation whose severity is diluted by pristine regions, whereas VisualQuality-R1~\cite{wu2025visualquality} assigns a relatively high holistic score despite prominent local defects. Such global assessments may obscure perceptually important artifacts and provide limited guidance for distortion-aware restoration~\cite{zhang2018ffdnet, zhang2020deep, jinjin2020pipal} and region-adaptive enhancement~\cite{moran2020deeplpf, guo2020zero}, which benefit from spatially grounded characterization of each defect and its severity.

Human observers, in contrast, rarely assess such complex degradations through an immediate monolithic judgment. Visual quality perception typically unfolds progressively: an initial \textit{glance} establishes a broad awareness of where potential degradations occur and what distortion patterns they resemble, after which closer \textit{scrutiny} examines the relevant local evidence to determine how severely visual quality is affected. This perceptual progression naturally gives rise to three interdependent questions: \textit{where} does the degradation occur, \textit{what} type of distortion is present, and \textit{how} severe is it? Crucially, these questions are neither independent nor equally difficult. Reliable severity estimation presupposes that the model has first attended to the correct region and identified the corresponding distortion pattern; otherwise, the severity signal may be dominated by irrelevant or pristine content.

Motivated by this observation, we propose GS-IQA, a unified framework that performs fine-grained image quality assessment through progressive \textit{Where--What--How} distortion reasoning. Rather than condensing heterogeneous degradations into a single global judgment, GS-IQA associates each degraded region with its distortion category and ordinal severity, thereby providing spatially grounded and verifiable quality interpretations. To learn these fine-grained associations, we develop a two-stage reinforcement learning paradigm based on Group Relative Policy Optimization (GRPO)~\cite{shao2024deepseekmath}, through which the model progresses from broad distortion perception to subtle severity discrimination. During the \textit{glance} stage, a perception-gated reward coordinates distortion localization, recognition, and severity learning according to their prerequisite relationships. In particular, the severity reward is activated only when the predicted region and distortion category are sufficiently reliable, preventing erroneous grounding or recognition from introducing misleading supervision while retaining coarse intensity cues for valid predictions. Having established robust \textit{Where--What} perception, the scrutiny stage shifts the optimization focus toward subtle severity variations that remain difficult to distinguish. Since repeatedly training on a fixed corpus can lead to early saturation, we introduce online reward-conditioned degradation generation, where category-wise severity statistics adapt the sampling probabilities of different distortions. By allocating more synthesized samples to underperforming categories, the training distribution evolves with the model, sustaining informative optimization and progressively sharpening fine-grained severity discrimination.
To support the training and systematic evaluation of fine-grained quality assessment, we further construct Diag-Bench, a region-level benchmark comprising 25K carefully curated samples across 12 distortion types and five ordinal severity levels. 

In summary, our contributions are as follows:
\begin{itemize}
    \item We introduce GS-IQA, a unified framework that reformulates fine-grained IQA as progressive \textit{Where--What--How} reasoning, explicitly associating each degraded region with its distortion type and ordinal severity.
    \item A \textit{Glance-to-Scrutiny} reinforcement learning paradigm is developed that respects the perceptual dependencies among distortion localization, recognition, and severity estimation. A perception-gated reward first establishes reliable \textit{Where--What} perception progressively refining fine-grained \textit{How} discrimination.
    \item We propose online reward-conditioned degradation generation for the scrutiny stage, which tracks category-wise severity performance and adaptively reallocates synthesis toward underperforming distortions, mitigating the saturation of severity learning on static data.
    \item Diag-Bench provides a region-level quality diagnosis benchmark comprising 25K samples across 12 distortion categories and five ordinal severity levels. Extensive experiments demonstrate the effectiveness of GS-IQA in fine-grained distortion assessment and its transferability to conventional global quality prediction.
\end{itemize}

\section{Related Works}
\subsection{Image Quality Assessment}
\textbf{Score-based Methods.}
Image quality assessment (IQA) is commonly studied under full-reference (FR) and no-reference (NR) settings.
FR-IQA compares distorted images against pristine references using structural similarity or learned perceptual representations~\cite{wang2004image, zhang2018unreasonable, prashnani2018pieapp, cao2022incorporating}, whereas NR-IQA predicts perceptual quality directly from distorted inputs, evolving from natural scene statistics~\cite{mittal2012no} to deep neural predictors~\cite{talebi2018nima, su2020blindly} and transformer-based architectures with multi-scale modeling~\cite{ke2021musiq}.
Despite strong correlation with human opinion scores, these methods largely compress visual quality into a single scalar, providing limited evidence about the spatial extent, distortion type, or severity underlying the prediction.

\textbf{MLLM-based Methods.}
MLLMs extend IQA beyond scalar prediction toward language-based quality understanding, covering quality scoring, descriptive assessment, and perceptual reasoning~\cite{you2024depicting, wu2023q, you2025teaching, wu2024q, wu2024towards, zhang2025teaching, zhang2025q}.
Recent studies further introduce reinforcement learning to improve perceptual alignment and distortion understanding~\cite{li2025qinsight, wu2025visualquality}.
Meanwhile, spatially grounded IQA methods associate quality judgments with localized visual evidence~\cite{chen2024q, chen2026grounding, peng2026iqa}.
SEAGULL~\cite{chen2024seagull} extends such analysis to distortion type and ordinal severity for specified regions, while Refine-IQA~\cite{jia2025refine} incorporates distortion recognition, severity estimation, and grounding into reinforcement fine-tuning.
These advances improve local quality understanding, yet the dependency among localization, recognition, and severity remains largely unexploited in learning. GS-IQA addresses this gap through progressive \emph{Where--What--How} reasoning, using reliable Where--What perception to constrain ordinal How supervision and thereby avoid misleading severity feedback from incorrect region--type associations. It further couples reward-conditioned online generation with the model's evolving severity weakness, enabling training to progress beyond coarse degradation perception toward fine-grained severity discrimination.

\subsection{Reinforcement Learning for MLLMs}
Reinforcement learning with verifiable rewards has proven effective for eliciting reasoning without dense annotation~\cite{shao2024deepseekmath, guo2025deepseek}, and has been extended to multimodal perception such as detection and grounding~\cite{liu2025visual, liu2025seg, bai2025univg}, where rule-based signals like IoU provide objective supervision.
In IQA, recent studies have introduced reward optimization for quality scoring and distortion understanding~\cite{li2025qinsight, wu2025visualquality, jia2025refine}, yet joint region-level diagnosis poses two additional challenges: independently optimized objectives can assign misleading severity credit to incorrectly localized or recognized regions, while fixed training corpora gradually lose informativeness as familiar degradations dominate.
Existing curriculum or reweighting strategies~\cite{bai2025univg, yan2026start}, which only reorganize available samples, cannot expand the training distribution. GS-IQA addresses both by gating severity feedback on verified localization and recognition, while using reward-conditioned online synthesis to replenish training with degradations from categories that remain difficult to grade.



\begin{figure*}[t]
\centering
\includegraphics[width=1.0\textwidth]{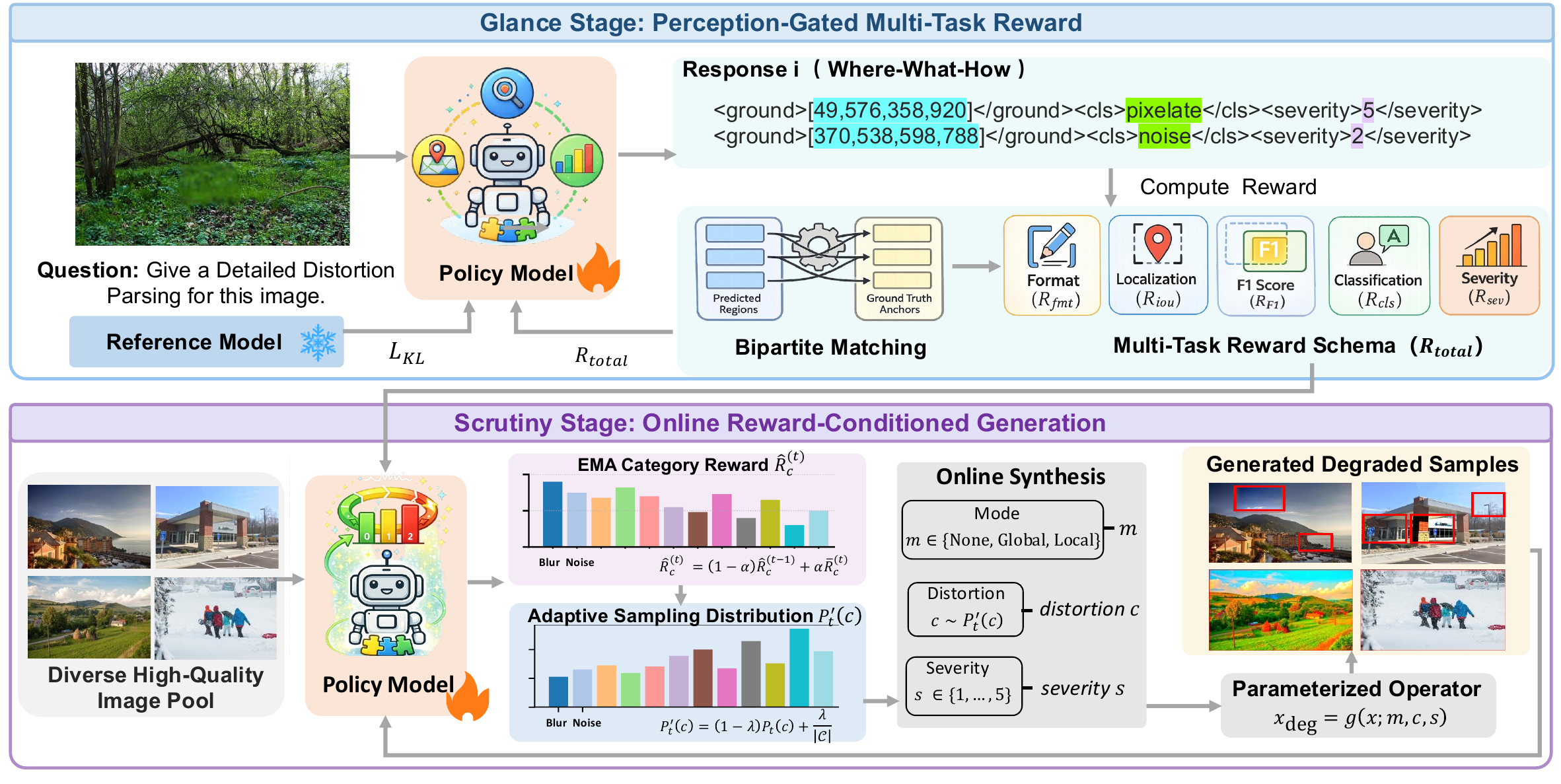}
\vspace{-5mm}
\caption{Framework of GS-IQA, which performs quality diagnosis by predicting triplets that localize degraded areas, identify distortion types, and estimate ordinal severity. Training follows a GRPO-based \textit{from-glance-to-scrutiny} scheme: the glance stage learns region-level grounding and recognition under perception-gated multi-task reward, while the scrutiny stage refines severity estimation with online reward-conditioned degradation generation of controllable hard examples.}
\vspace{-3mm}
\label{fig:framework}
\end{figure*}

\section{Methodology}

\subsection{Overview of GS-IQA}
As illustrated in Figure~\ref{fig:framework}, GS-IQA formulates region-level image quality assessment as progressive \textit{Where--What--How} diagnosis: localizing degraded regions, recognizing their distortion types, and estimating their ordinal severity levels. These dimensions are interdependent: since the same ordinal level carries different perceptual salience across distortion types, a severity prediction is meaningful only once the region and its type are correct.
GS-IQA therefore adopts a two-stage GRPO framework, in which the glance stage establishes region-level perception through perception-gated multi-task learning, activating severity feedback only for spatially matched and correctly recognized regions. The scrutiny stage then tracks category-wise severity performance and adaptively emphasizes underperforming distortions through online severity-controllable synthesis, further refining How discrimination.
\subsection{Glance Stage: Perception-Gated Multi-Task Learning}
Building on the MLLM's low-level visual priors, the Glance stage learns region-level \textit{Where--What--How} perception under GRPO. A perception-gated multi-task reward jointly supervises distortion localization and category recognition, while activating ordinal severity feedback only for spatially matched regions with correctly recognized distortion types. This design provides reliable supervision for localized degradation perception and establishes a robust perceptual foundation for the subsequent Scrutiny stage.

\textbf{Format Reward.}
To enable reliable parsing and automatic reward computation, we adopt a lightweight serialization convention for the predicted \textit{Where--What--How} associations. Specifically, the reasoning process is delimited by $\texttt{<think>}...\texttt{</think>}$, followed by the final prediction within $\texttt{<answer>}...\texttt{</answer>}$. Within the answer span, each detected region is represented by a region--category--severity association $\hat{\mathbf{t}}_i=(\hat{\mathbf{b}}_i,\hat{c}_i,\hat{s}_i)$, serialized as:
\begin{equation}
\hat{\mathbf{t}}_i \equiv
\texttt{<ground>}\hat{\mathbf{b}}_i\texttt{</ground>},
\texttt{<cls>}\hat{c}_i\texttt{</cls>},
\texttt{<severity>}\hat{s}_i\texttt{</severity>},
\end{equation}
where $\hat{\mathbf{b}}_i=[\hat{x}_{i1},\hat{y}_{i1},\hat{x}_{i2},\hat{y}_{i2}]$, $\hat{c}_i$, and $\hat{s}_i$ denote the predicted bounding box, distortion category, and severity level, respectively. The associations extracted from the answer constitute the prediction set $\hat{\mathcal{T}} = \{\hat{t}_i\}_{i=1}^N$, which provides the common input for the grounding and severity rewards below.
We assign $R_{\mathrm{fmt}}=1$ when the reasoning and answer spans are correctly delimited and the answer contains at least one well-formed association, and $R_{\mathrm{fmt}}=0$ otherwise. This reward only ensures parseability, while semantic correctness is evaluated by the subsequent perception rewards.
For pristine images, the canonical association $([0,0,1000,1000], \texttt{None}, 0)$ is used, allowing them to share the same structured output and reward pipeline as degraded images.


\textbf{Degradation Grounding Reward.}
This reward supervises degradation grounding, covering spatial localization (\textit{Where}) and distortion identification (\textit{What}). Since a single image may contain multiple degraded regions and co-occurring artifacts, the predictions form an unordered set rather than an ordered sequence. To handle this permutation ambiguity, we establish an optimal bipartite matching between predicted and ground-truth triplets before reward computation, ensuring each prediction is compared to the best-aligned ground-truth region.

Specifically, we formulate this correspondence as a bipartite assignment problem. 
Let $\hat{\mathcal{T}} = \{\hat{\textbf{t}}_i\}_{i=1}^N$ and $\mathcal{T} = \{\textbf{t}_j\}_{j=1}^M$ denote the predicted and ground-truth triplet sets, respectively. Each triplet $\textbf{t}=(\textbf{b},c,s)$ comprises a bounding box $\textbf{b}$, a distortion category $c$, and an ordinal severity level $s$. To establish correspondence between the two sets, we define $\Phi(\hat{\mathbf{t}}_i,\mathbf{t}_j)$ as their pairwise perception score and seek the assignment that maximizes the overall compatibility:
\begin{align}
    \hat{\sigma} = \arg \max_{\sigma \in \Omega_{N,M}} \sum_{i=1}^{N} \Phi(\hat{\textbf{t}}_i, \textbf{t}_{\sigma(i)}),
\end{align}
where $\Omega_{N,M}$ denotes the set of valid one-to-one matchings from predictions to ground-truth, obtained using the Hungarian algorithm~\cite{kuhn1955hungarian}. Based on optimal assignment, we retain the spatially valid pairs whose intersection-over-union exceeds: $\mathcal{M}=\{(i,j)\mid j=\hat{\sigma}(i),\ \mathrm{IoU}(\hat{\textbf{b}}_i,\textbf{b}_j)\ge 0.5\}$. The localization and classification rewards are then computed over the matched pairs:
\begin{align}
    R_{\text{iou}} &= \frac{1}{|\mathcal{M}|} \sum_{(i, j) \in \mathcal{M}} \text{IoU}(\hat{\textbf{b}}_i, \textbf{b}_j),
    \quad R_{\text{cls}} = \frac{1}{|\mathcal{M}|} \sum_{(i, j) \in \mathcal{M}} \mathbb{I}(\hat{c}_i = c_j).
\end{align}

Both rewards are set to zero when no valid pair is found. To penalize both missed regions and hallucinated predictions, we incorporate the F1-style grounding reward $\frac{2|\mathcal M|}{N+M}$, which is equivalent to the harmonic mean of matching precision $|\mathcal M|/N$ and recall $|\mathcal M|/M$.

\textbf{Perception-Gated Severity Reward.}
Severity estimation is inherently category-dependent, as the same ordinal level may exhibit markedly different perceptual salience across distortion types (e.g., {Level 1} saturation can be more noticeable than {Level 1} blur). A severity prediction is therefore meaningful only relative to a correctly recognized distortion category. For the spatially matched pairs in $\mathcal{M}$, we activate severity supervision only when the predicted category agrees with the ground truth, conditioning \textit{How} discrimination on reliable \textit{Where–What} perception.


To provide denser feedback while preserving the ordinal structure of severity, we assign graded credit according to the distance between the predicted and ground-truth levels. For each matched pair $(i,j)\in\mathcal{M}$, we define the ordinal severity distance as
\begin{equation}
d^{\mathrm{sev}}_{ij}=|\hat{s}_i-s_j|.
\end{equation}
The tiered ordinal reward assigns progressively lower credit as the severity distance increases, taking
$r_{\mathrm{ord}}=1,\alpha_1,\alpha_2,0$ for $d=0,1,2,$ and $d\geq3$, respectively, where $\alpha_1>\alpha_2>0$ enforces monotonic decay and $(\alpha_1,\alpha_2)=(0.6,0.2)$ in practice.
Crucially, severity credit is released only when both spatial correspondence and distortion recognition are correct. We therefore define
\begin{equation}
R_{\mathrm{sev}}
=
\frac{1}{|\mathcal{M}|}
\sum_{(i,j)\in\mathcal{M}}
\mathbb{I}(\hat{c}_i=c_j)\,
r_{\mathrm{ord}}\!\left(d^{\mathrm{sev}}_{ij}\right),
\end{equation}
Unmatched or misclassified regions contribute zero reward. This normalization prevents selective omission of difficult regions from inflating the severity reward, while preserving the prerequisite relationship that valid \emph{How} supervision requires reliable \emph{Where--What} perception.


\textbf{Overall Multi-Task Reward.}
The above components are combined to jointly optimize format validity, degradation grounding, and severity perception:
\begin{equation}
    R_{\text{total}} = R_{\text{fmt}} + R_{\text{iou}} + R_{\text{F1}} + R_{\text{cls}} + R_{\text{sev}}.
\end{equation}

\subsection{Scrutiny Stage: Online Reward-Conditioned Generation}
After the \textit{Glance} stage establishes reliable region-level degradation perception, further improvement in severity discrimination is often limited by the fixed composition of static training data. A fixed corpus provides finite combinations of distortion patterns and severity appearances, causing optimization to gradually concentrate on already well-learned degradations. To sustain informative training, the Scrutiny stage introduces online reward-conditioned degradation generation, which tracks category-wise severity performance and adaptively reallocates sampling toward distortions that remain difficult to grade. As the policy evolves, the training distribution evolves accordingly, continually refreshing fine-grained supervision around the model's current weaknesses.

\textbf{Reward-Driven Distribution Evolution.}
We maintain an online profile of category-wise severity performance to guide degradation sampling.
At step $t$, for each distortion category $c$, we collect its spatially matched regions as
$\mathcal{B}^{(t)}_c=\{(i,j)\in\mathcal{M}\mid c_j=c\}$.
The batch-level score measures severity performance over these regions, where category-mismatched predictions receive zero credit. To reduce batch-wise fluctuations, we further maintain an exponential moving average (EMA):
\begin{equation}
\bar{R}^{(t)}_c =
\frac{1}{|\mathcal{B}^{(t)}_c|}
\sum_{(i,j)\in\mathcal{B}^{(t)}_c}
\mathbb{I}(\hat{c}_i=c_j)\,
r_{\mathrm{ord}}\!\left(d^{\mathrm{sev}}_{ij}\right), \quad
\hat{R}^{(t)}_c =
(1-\alpha)\hat{R}^{(t-1)}_c
+\alpha\bar{R}^{(t)}_c ,
\end{equation}
where $\bar{R}^{(t)}_c$ reflects the current batch-wise severity performance of category $c$, and $\hat{R}^{(t)}_c$ provides a smoothed estimate of its recent learning status. If category $c$ is absent from the current batch, its EMA statistic remains unchanged. A lower $\hat{R}^{(t)}_c$ therefore indicates a distortion whose severity remains harder to grade. We convert these statistics into an adaptive sampling distribution by applying a temperature-scaled softmax over the negative rewards:
\begin{equation}
P_t(c)=\frac{\exp\!\left(-\hat{R}^{(t)}_c/\tau\right)}
{\sum_{c'\in\mathcal{C}}
\exp\!\left(-\hat{R}^{(t)}_{c'}/\tau\right)},
\quad
P'_t(c)=(1-\lambda)P_t(c)+\frac{\lambda}{|\mathcal{C}|},
\end{equation}
where $\tau$ controls how strongly sampling concentrates on low-reward categories, while $\lambda$ preserves exploration through a uniform prior. Consequently, categories with persistently weak severity performance receive more training samples, whereas well-learned categories gradually recede.
Unlike reweighting a fixed corpus, this distribution directly governs online degradation synthesis, allowing the training data to evolve with the policy.

\textbf{Online Degradation Synthesis.}
Given the adaptive sampling distribution $P'_t(c)$, we synthesize training samples online by sampling a degradation mode $m \in \{\texttt{None},\texttt{Global},\texttt{Local}\}$, a distortion category $c \sim P'_t(c)$, and an ordinal severity level $s \in \{1, \dots, 5\}$. The mode determines whether we keep the image pristine, apply a single global degradation, or inject multiple localized degradations into randomly sampled patches. The generator $g(\cdot)$ applies a degradation operator to a pristine image $x$ to produce a degraded image:
\begin{align}
    x_{\text{deg}} = g(x; m, c, s).
\end{align}

Crucially, it emits structured supervision together with the synthesized image. For each degraded region $k$, we record its bounding box $b_k$, category $c_k$, and severity level $s_k$, forming a set-structured annotation $\mathcal{T} = \{(b_k, c_k, s_k)\}_{k=1}^{K}$. In the local mode, the set of severity $\left\{s_k\right\}$ is sampled independently for each injected region, enabling multiple localized degradations with controllable severity within one image. By combining reward-driven category sampling with controllable severity synthesis, this stage continually expands fine-grained supervision while concentrating training capacity on distortions with weak severity estimation.


\subsection{Diag-Bench: Benchmark for Fine-grained Quality Diagnosis}

We construct {Diag-Bench}, a 25K-sample benchmark that formulates IQA as a fine-grained \textit{Where--What--How} diagnosis, requiring models to localize degraded regions, identify distortion types, and estimate ordinal severity. Unlike existing benchmarks dominated by image-level scores or coarse descriptions, Diag-Bench combines authentic and synthetic degradations with region-level annotations. Following prior IQA studies~\cite{depictqa_v2}, we consider 12 representative distortion categories---noise, blur, compression, overexposure, underexposure, high contrast, low contrast, oversaturate, desaturate, oversharpen, pixelate, quantization---each with five ordinal severity levels.

The authentic subset is built upon ViDA-UGC~\cite{liao2025vida}, a large-scale UGC dataset with region-level distortion annotations, which we semantically map onto our predefined categories while discarding those without a counterpart. To ensure unambiguous severity annotation, we retain only images carrying a single distortion category and annotate their ordinal severity, yielding 3,470 training and 868 testing samples.
The synthetic subset provides scalable and controllable supervision over distortion type, spatial extent, and severity. We sample 8,000 pristine images from KADIS-700K~\cite{lin2019kadid} and apply a two-stage quality screening procedure: Q-Insight~\cite{li2025qinsight} and VisualQuality-R1~\citep{wu2025visualquality} first filter sources with potential pre-existing degradations, followed by manual inspection to remove residual low-quality images, leaving 4,275 high-quality sources. To avoid source-level leakage, we partition the 4,275 pristine images into 3,775 training and 500 testing sources before any degradation is applied. All static samples and online degradations used in both the Glance and Scrutiny stages are generated exclusively from the training-source pool, while the test sources remain completely unseen throughout optimization. Parameterized OpenCV operators~\citep{depictqa_v2,opencv_library,7752930,lin2019kadid} generate pristine, global, and non-overlapping local degradations in a 1:2:2 ratio, with 1--3 degraded regions per local sample. Each image is annotated with triplets $\{(b_k,c_k,s_k)\}_{k=1}^{K}$, where $s_k\in\{1,\ldots,5\}$ denotes ordinal severity; pristine images use the canonical annotation $([0,0,1000,1000],\texttt{None},0)$. To ensure the perceptibility of the synthesized label, human experts review all candidates and discard ambiguous ones, retaining 21,275 samples, 18,775 for training and 2,500 for testing.

\section{Experiments}
\subsection{Experimental Setup}
\textbf{Implementation Details.}
We adopt Qwen3-VL-4B~\cite{Qwen3-VL} as the base model and optimize it through two consecutive reinforcement stages without preceding supervised fine-tuning. The Glance stage is trained on the static training split of Diag-Bench, whereas the Scrutiny stage employs online reward-conditioned degradation generation for severity refinement. GRPO samples a group of $G=5$ responses per prompt with a global batch size of 128. In the scrutiny stage, the online sampler tracks per-category severity performance with EMA momentum $\alpha=0.1$, using temperature $\tau=1.0$ for category reweighting, and mixes the sampling distribution with a uniform prior using $\lambda=0.05$. All experiments are conducted with NVIDIA H800 GPUs.

\textbf{Evaluation Protocols and Metrics.}
We evaluate GS-IQA under three complementary settings covering core diagnostic capability, cross-dataset generalization, and transferability to holistic quality assessment. On Diag-Bench, region-level evaluation assesses the complete \textit{Where--What--How} diagnosis: localization is measured by precision, recall, F1-score, and mIoU, while distortion recognition and severity estimation are measured by classification accuracy (ClsAcc) and severity accuracy (SevAcc), respectively. Cross-dataset generalization is evaluated on established external benchmarks without dataset-specific fine-tuning, where ClsAcc and SevAcc are reported under the global-degradation setting. Finally, we assess the transferability of \textit{Where--What--How} learning to global IQA, with Spearman Rank-Order Correlation Coefficient (SRCC) and Pearson Linear Correlation Coefficient (PLCC) measuring how effectively region-aware distortion semantics and ordinal severity cues support holistic quality prediction.

\begin{table*}[t]
\centering
\caption{Fine-grained image quality diagnosis results on Diag-Bench. We evaluate distortion localization (Where), type recognition (What), and severity estimation (How). \textsuperscript{\dag} denotes task-aligned variants. The best and second-best results are marked in \textbf{bold} and \underline{underlined}, respectively.}
\vspace{-3mm}
\footnotesize
\label{tab:main_results}
\setlength{\tabcolsep}{3mm}
\resizebox{\textwidth}{!}{
\begin{tabular}{lcccccc}
\toprule
\multirow{2}{*}{{Method}}
& \multicolumn{4}{c}{{Where}}
& {What}
& {How} \\ \cmidrule(lr){2-5} \cmidrule(lr){6-6} \cmidrule(lr){7-7}  
& {Precision}
& {Recall}
& {F1}
& {mIoU}
& {ClsAcc}
& {SevAcc} \\
\midrule
\rowcolor{gray!12}
\multicolumn{7}{c}{\textit{General-purpose MLLMs}} \\
Gemini-2.5-Pro~\cite{comanici2025gemini}
& 0.343 & 0.428 & 0.381 & 0.920 & 0.466 & 0.363 \\
Gemini-3-Pro
& 0.722 & 0.563 & 0.633 & 0.921 & 0.603 & 0.563 \\
GPT-4o~\cite{hurst2024gpt}
& 0.183 & 0.227 & 0.202 & 0.876 & 0.600 & 0.600 \\
GPT-5~\cite{singh2025openai}
& 0.138 & 0.269 & 0.183 & 0.854 & 0.677 & 0.400 \\
Qwen3-VL-4B~\cite{Qwen3-VL} (Base)
& 0.544 & 0.176 & 0.266 & 0.905 & \underline{0.931} & 0.316 \\
Qwen3-VL-235B~\cite{Qwen3-VL}
& 0.624 & 0.330 & 0.431 & \underline{0.954} & 0.535 & 0.677 \\
\rowcolor{gray!12}
\multicolumn{7}{c}{\textit{Grounding-oriented Models}} \\
Shikra-7B~\cite{chen2023shikra}
& 0.019 & 0.184 & 0.034 & 0.602 & 0.041 & N/A \\

Ferret-7B~\cite{you2023ferret}
& 0.032 & 0.286 & 0.058 & 0.687 & 0.055 & N/A \\

Kosmos-2-1.6B~\cite{peng2023kosmos}
& 0.012 & 0.125 & 0.022 & 0.583 & 0.033 & N/A \\

GroundingGPT-7B~\cite{li2024groundinggpt}
& 0.024 & 0.203 & 0.043 & 0.653 & 0.067 & N/A \\

\rowcolor{gray!12}
\multicolumn{7}{c}{\textit{IQA-specific Models}} \\


Q-Insight~\cite{li2025qinsight}
& N/A & N/A & N/A & N/A & 0.346 & 0.502 \\

VisualQuality-R1~\cite{wu2025visualquality}
& N/A & N/A & N/A & N/A & 0.284 & 0.606 \\

\rowcolor{gray!12}
\multicolumn{7}{c}{\textit{Fine-grained IQA Models}} \\
Supervised Fine-tuning
& 0.854 & \underline{0.861} & 0.857 & 0.926 & 0.467 & 0.585 \\


Q-Insight\textsuperscript{\dag}~\cite{li2025qinsight}
& 0.889 & 0.814 & 0.850 & 0.938 & 0.892 & 0.652\\


VisualQuality-R1\textsuperscript{\dag}~\cite{wu2025visualquality}
& 0.910 & 0.858 & 0.883 & 0.946 & 0.907 & 0.438 \\


\rowcolor{blue!10}
\textbf{GS-IQA-Glance (Ours)}
& \underline{0.948}
& 0.848
& \underline{0.895}
& 0.953
& 0.921
& \underline{0.678} \\

\rowcolor{blue!10}
\textbf{GS-IQA-Scrutiny (Ours)}
& \textbf{0.953}
& \textbf{0.909}
& \textbf{0.930}
& \textbf{0.967}
& \textbf{0.947}
& \textbf{0.732} \\

\bottomrule
\end{tabular}
}
\vspace{-7mm}
\end{table*}

\subsection{Comparison with State-of-the-Art Methods}
\textbf{Fine-Grained Quality Diagnosis.}
Table~\ref{tab:main_results} compares GS-IQA with general-purpose MLLMs, grounding models, IQA-specific models, and task-aligned fine-grained baselines on Diag-Bench. General-purpose and grounding-oriented models show limited transfer to low-level degradation localization: GPT-4o and GPT-5 achieve F1 scores of only 0.202 and 0.183, while conventional grounding models remain below 0.06. This reflects the difficulty of localizing degradations, which are often weakly bounded and spatially diffuse rather than semantically well-defined objects.

\begin{wraptable}{r}{0.39\textwidth}
    \centering
    \vspace{-8pt}
    \caption{Global degradation diagnosis.}
    \label{tab:global_diagnosis}
    \resizebox{0.39\textwidth}{!}{
    \begin{tabular}{lcc}
        \toprule
        Method & ClsAcc & SevAcc \\
        \midrule
        AgenticIR              & 0.426 & 0.237 \\
        Q-Insight              & 0.434 & 0.365 \\
        VisualQuality-R1       & 0.482 & 0.359 \\
        \textbf{GS-IQA-Scrutiny} & \textbf{0.916} & \textbf{0.735} \\
        \bottomrule
    \end{tabular}}
    \vspace{-8pt}
\end{wraptable}
IQA-specific MLLMs provide stronger quality awareness, but their holistic formulations neither ground individual degraded regions nor reliably disentangle region-wise distortion attributes. Supervised fine-tuning achieves competitive localization but substantially weaker recognition and severity estimation. To control for output-format differences, we adapt Q-Insight and VisualQuality-R1 to the same prompts, training data, and structured prediction format, while preserving their respective optimization objectives. Both task-aligned variants improve markedly over SFT in localization and recognition; however, neither explicitly exploits ordinal severity distance during optimization, with Q-Insight$^\dagger$ reaching a SevAcc of 0.652 and VisualQuality-R1$^\dagger$ 0.438.
GS-IQA-Glance further raises SevAcc to 0.678 through perception-gated ordinal rewards, while GS-IQA-Scrutiny uses category-wise severity feedback to adapt online degradation sampling, increasing SevAcc to 0.732 while also improving Recall from 0.848 to 0.909, F1 from 0.895 to 0.930, and ClsAcc from 0.921 to 0.947. These results demonstrate that explicitly modeling the dependencies among localization, recognition, and severity is effective for joint fine-grained quality diagnosis.

\textbf{Global Degradation Diagnosis.}
We further evaluate the diagnostic capability of GS-IQA beyond localization on globally degraded images, where each image contains a single distortion with uniform severity. This setting removes the grounding requirement and isolates the \textit{What} and \textit{How} capabilities. As shown in Table~\ref{tab:global_diagnosis}, GS-IQA-Scrutiny substantially outperforms existing IQA methods, improving the best competing ClsAcc from 0.482 to 0.916 and SevAcc from 0.365 to 0.735. These results demonstrate that the advantage of GS-IQA extends beyond spatial grounding to distortion recognition and fine-grained severity discrimination. 

\begin{table}[t]
\centering
\caption{Cross-dataset generalization of distortion diagnosis on external IQA benchmarks without dataset-specific fine-tuning. Best results are highlighted in \textbf{bold}.}
\vspace{-2mm}
\label{tab:cross_dataset_generalization}
\resizebox{0.8\textwidth}{!}{
\begin{tabular}{l|l|ccccc}
\toprule
{Method} & {Metric} & {DQ-495K} & {KADID-10k} & {TID2013}  & Waterloo \\
\midrule

\multirow{2}{*}{Q-Insight~\cite{li2025qinsight}}
& ClsAcc & 0.284 & 0.242 & 0.443  & 0.504  \\
& SevAcc & 0.398 & 0.430 & 0.085  & 0.254  \\
\midrule

\multirow{2}{*}{VisualQuality-R1~\cite{wu2025visualquality}}
& ClsAcc & 0.157  & 0.188 & 0.158 & 0.332 \\
& SevAcc & 0.498 & 0.232& 0.306  & 0.074 \\
\midrule
\multirow{2}{*}{\textbf{GS-IQA (Ours)}}
& ClsAcc & \textbf{0.787} & \textbf{0.658} & \textbf{0.736} & \textbf{0.832} \\
& SevAcc & \textbf{0.532} & \textbf{0.514} & \textbf{0.354} & \textbf{0.399} \\
\bottomrule
\end{tabular}}
\vspace{-4mm}
\end{table}

\begin{table}[t]
\centering
\caption{Holistic quality prediction transferability across five external IQA benchmarks.}
\vspace{-2mm}
\label{tab:global_iqa}
\setlength{\tabcolsep}{4.0pt}
\renewcommand{\arraystretch}{1.08}
\resizebox{\textwidth}{!}{
\begin{tabular}{l cc cc cc cc cc  cc}
\toprule
\multirow{2}{*}{{Method}}
& \multicolumn{2}{c}{{BID}}
& \multicolumn{2}{c}{{CLIVE}}
& \multicolumn{2}{c}{{SPAQ}}
& \multicolumn{2}{c}{{KonIQ}}
& \multicolumn{2}{c}{{AGIQA}}
& \multicolumn{2}{c}{{AVG.}} \\
\cmidrule(lr){2-3}
\cmidrule(lr){4-5}
\cmidrule(lr){6-7}
\cmidrule(lr){8-9}
\cmidrule(lr){10-11}
\cmidrule(lr){12-13}
& SRCC & PLCC
& SRCC & PLCC
& SRCC & PLCC
& SRCC & PLCC
& SRCC & PLCC
& SRCC & PLCC \\
\midrule

Q-Align~\cite{wu2023q}
& 0.576 & 0.651
& 0.554 & 0.643
& 0.767 & 0.779
& 0.573 & 0.612
& 0.682 & 0.705
& 0.630 & 0.678 \\

DeQA-Score~\cite{you2025teaching}
& 0.702 & 0.743
& 0.743 & 0.795
& 0.852 & 0.858
& 0.677 & 0.703
& {0.738} & \underline{0.790}
& 0.742 & 0.778 \\

Q-Insight~\cite{li2025qinsight}
& 0.806 & 0.818
& 0.804 & 0.837
& \underline{0.907} & \underline{0.912}
& 0.812 & 0.809
& 0.657 & 0.705
& 0.797 & 0.816 \\

VisualQuality-R1~\cite{wu2025visualquality}
& \underline{0.811} & \underline{0.820}
& \underline{0.811} & \underline{0.844}
& \textbf{0.913} & \textbf{0.917}
& \underline{0.855} & \underline{0.870}
& \textbf{0.754} & \textbf{0.820}
& \underline{0.829} & \underline{0.854} \\

\rowcolor{blue!10}
\textbf{GS-IQA (Ours)}
& \textbf{0.886} & \textbf{0.884}
& \textbf{0.893} & \textbf{0.898}
& 0.886 & 0.882
& \textbf{0.885} & \textbf{0.901}
& \underline{0.749} & 0.780
& \textbf{0.860} & \textbf{0.869} \\

\bottomrule
\end{tabular}
}
\vspace{-6mm}
\end{table}

\begin{table}[t]
\centering
\caption{{Ablation studies of GS-IQA}, including (a) the perception-gated severity reward in the initial glance stage, and (b) reward-conditioned online generation in the scrutiny stage.}
\vspace{-2mm}
\label{tab:unified_ablation}
\setlength{\tabcolsep}{4mm}
\resizebox{\columnwidth}{!}{
\begin{tabular}{lcccccc}
\toprule
\textbf{Configuration} & \textbf{Precision} & \textbf{Recall} & \textbf{F1} & \textbf{mIoU} & \textbf{ClsAcc} & \textbf{SevAcc} \\ \midrule
\rowcolor[HTML]{F2F2F2} \multicolumn{7}{c}{\textit{(a) Severity Reward in Glance Stage}} \\ 
Hard Binary Reward & 0.955 & 0.800 & 0.871 & 0.960 & 0.939 & 0.652 \\
Continuous Linear Reward & 0.954 & 0.879 & 0.915 & 0.953 & 0.926 & 0.627 \\
Ordinal Ranking Reward & 0.966 & 0.872 & 0.916 & 0.960 & 0.924 & 0.304 \\
\textbf{Perception-gated Reward} & \textbf{0.948} & \textbf{0.848} & \textbf{0.895} & \textbf{0.953} & \textbf{0.921} & \textbf{0.678} \\ 
\quad w/o Perception Gating & 0.950 & 0.816 & 0.878 & 0.957 & 0.927 & 0.651 \\ \midrule 
\rowcolor[HTML]{F2F2F2} \multicolumn{7}{c}{\textit{(b) Online Generation in Scrutiny Stage}} \\ 
Static Data Only & 0.941 & 0.849 & 0.893 & 0.957 & 0.945 & 0.682 \\
Random Online Generation & 0.967 & 0.897 & 0.931 & 0.967 & 0.947 & 0.694 \\
\textbf{Reward-conditioned Generation} & \textbf{0.953} & \textbf{0.909} & \textbf{0.930} & \textbf{0.967} & \textbf{0.947} & \textbf{0.732} \\
\quad w/o glance initialization & 0.957 & 0.854 & 0.902 & 0.960 & 0.914 & 0.665 \\ \bottomrule
\end{tabular}
}
\vspace{-3mm}
\end{table}

\subsection{Generalization and Transferability}
\textbf{Cross-Dataset Diagnostic Generalization.}
We evaluate the transferability of the learned distortion semantics and severity cues beyond Diag-Bench on DQ-495K, KADID-10K, TID2013, and Waterloo, without dataset-specific fine-tuning. Since these benchmarks provide image-level annotations, evaluation focuses on the transferable \textit{What} and \textit{How} dimensions using the distortion categories and five-level severity labels that can be mapped to our taxonomy. As shown in Table~\ref{tab:cross_dataset_generalization}, GS-IQA consistently outperforms Q-Insight and VisualQuality-R1 across all four datasets, improving ClsAcc over the strongest baseline by 50.3, 41.6, 29.3, and 32.8 points, respectively, while achieving gains of up to 14.5 points in SevAcc. These consistent improvements demonstrate that GS-IQA learns transferable distortion and severity representations that generalize beyond the content distribution and synthesis patterns of Diag-Bench.

\textbf{Holistic Quality Prediction Transferability.}
We further evaluate the transferability of the fine-grained representations learned through \textit{Where--What--How} diagnosis to conventional global IQA. Following the training protocol of VisualQuality-R1, we initialize a global quality regressor from GS-IQA and fine-tune it using the same training data, dataset splits, and regression objective. As shown in Table~\ref{tab:global_iqa}, GS-IQA achieves the best average performance across five external benchmarks, reaching 0.860 SRCC and 0.869 PLCC. Compared with VisualQuality-R1, the average SRCC and PLCC improve by 0.031 and 0.015, respectively, with GS-IQA ranking first on BID, CLIVE, and KonIQ. 
These results indicate that Where--What--How learning captures transferable perceptual representations that can be effectively aggregated into calibrated image-level scores. GS-IQA therefore supports both interpretable local diagnosis and conventional global quality prediction.

\subsection{Ablation Studies}

\textbf{Impact of Severity Reward.}
Table~\ref{tab:unified_ablation}(a) examines how different reward formulations affect severity learning in the Glance stage. The proposed perception-gated reward achieves the highest SevAcc of 0.678, improving over the hard binary reward (0.652) while maintaining competitive localization performance. Continuous linear and ordinal ranking rewards (detailed in Supplementary materials) retain strong F1 scores but yield noticeably lower SevAcc of 0.627 and 0.304, respectively, indicating that reward design plays a critical role in preserving exact severity calibration rather than merely relative intensity ordering. Moreover, removing perception gating reduces both F1 (0.895$\rightarrow$0.878) and SevAcc (0.678$\rightarrow$0.651), supporting our key design that severity feedback should be conditioned on reliable spatial and categorical perception. Together, these results show that effective fine-grained severity learning requires both ordinally informative feedback and valid \textit{Where--What} associations.


\textbf{Impact of Online Generation.}
Table~\ref{tab:unified_ablation}(b) evaluates various data strategies in the scrutiny stage, revealing that training exclusively on static data leads to a performance bottleneck, whereas online generation effectively overcomes this limit by introducing dynamic perceptual challenges. Among online strategies, our reward conditioned generation achieves a higher SevAcc of 0.732 compared to random generation. The higher SevAcc of reward-conditioned generation indicates that category-wise severity feedback provides a more informative sampling signal than random online synthesis, allocating additional training to distortions that remain difficult to grade. Removing Glance initialization consistently degrades localization, recognition, and severity performance, confirming that the static perception stage provides a stronger initialization for subsequent adaptive refinement.

\textbf{Sample Efficiency of Generative Scrutiny.}
Table~\ref{tab:sample_efficiency}(a) investigates the sample efficiency of reward-conditioned generation under different source-pool sizes. Notably, with only 50 source images (1\% of the training pool), GS-IQA already achieves an F1 score of 0.902, outperforming the Glance-stage baseline trained on static data. As the source pool expands from 50 to 3,775 images, performance improves steadily from 0.902 to 0.930 in F1 and from 0.689 to 0.732 in SevAcc. These results show that online degradation synthesis can effectively reuse limited pristine content to produce diverse region--category--severity supervision, while broader source diversity further benefits fine-grained severity learning. This makes the Scrutiny stage substantially less dependent on manually constructed training samples and enables scalable refinement from a compact pool of pristine images.


\begin{figure}[t]
    \centering
    \includegraphics[width=0.85\linewidth]{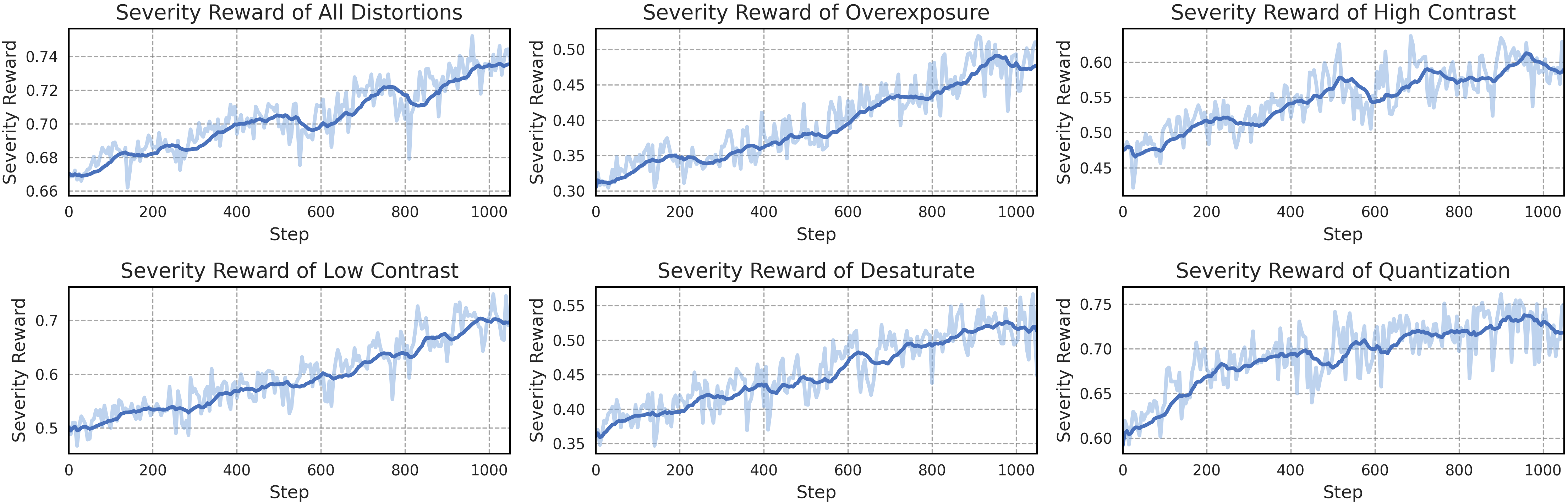}
    \vspace{-2mm}
    \caption{Learning dynamics of severity perceptual refinement in GS-IQA-Scrutiny.}
    \label{fig:dynamics}
    \vspace{-6mm}
\end{figure}


\begin{wraptable}{r}{0.38\columnwidth}
\vspace{-0.8\baselineskip}
\centering
\scriptsize
\setlength{\tabcolsep}{2mm}
\caption{Ablation on sample efficiency and model scaling.}
\label{tab:sample_efficiency}
\vspace{-0.5em}
\begin{tabular}{@{}lccc@{}}
\toprule
\textbf{Configuration} & \textbf{F1} & \textbf{ClsAcc} & \textbf{SevAcc} \\
\midrule
\rowcolor[HTML]{F7F7F7}
\multicolumn{4}{@{}c@{}}{\textit{(a) Sample Efficiency}} \\
50 (1\%)       & 0.902 & 0.923 & 0.689 \\
500 (13\%)     & 0.915 & 0.932 & 0.705 \\
1,000 (26\%)   & 0.922 & 0.939 & 0.718 \\
2,000 (53\%)   & 0.927 & 0.943 & 0.727 \\
\textbf{3,775 (100\%)}
               & \textbf{0.930} & \textbf{0.947} & \textbf{0.732} \\
\midrule
\rowcolor[HTML]{F7F7F7}
\multicolumn{4}{@{}c@{}}{\textit{(b) Model Capacity}} \\
GS-IQA-4B      & 0.930 & 0.947 & 0.732 \\
GS-IQA-8B      & 0.944 & 0.942 & 0.740 \\
\textbf{GS-IQA-32B}
               & \textbf{0.952} & \textbf{0.956} & \textbf{0.754} \\
\bottomrule
\end{tabular}
\vspace{-0.8\baselineskip}
\end{wraptable}
\textbf{Scaling Analysis of Model Capacity.}
Table~\ref{tab:sample_efficiency}(b) shows that increasing model capacity yields diminishing gains in localization but more consistent improvements in severity estimation. As the model scales from 4B to 32B, F1 increases from 0.930 to 0.952, while SevAcc improves from 0.732 to 0.754.
This trend suggests that fine-grained severity discrimination benefits more from increased model capacity than spatial localization, further highlighting severity estimation as the more challenging component of the \textit{Where--What--How} diagnosis.



\textbf{Dynamics of Perceptual Refinement.}
Figure~\ref{fig:dynamics} shows the evolution of severity rewards during the Scrutiny stage. The overall reward rises steadily, with consistent improvements observed for representative distortions such as overexposure, high contrast, low contrast, desaturation, and quantization. This behavior reflects the intended effect of reward-conditioned sampling: categories with weaker severity estimation continue to receive additional training exposure, enabling progressive refinement rather than early saturation on a fixed training distribution.

\section{Conclusion}
In this paper, we present GS-IQA, a fine-grained image quality diagnosis framework that reasons about degradations progressively, from an initial glance to closer scrutiny. Instead of collapsing heterogeneous degradations into a single score, GS-IQA characterizes each along the \textit{Where--What--How} dimensions--locating the affected region, recognizing its type, and grading its severity. As a severity judgment holds only where the region and its type are already correct, we cast these interdependent goals into a two-stage reinforcement learning scheme: perception-gated rewards first consolidate localization and recognition in the glance stage, and online reward-conditioned degradation generation then supplies targeted hard cases that sharpen severity estimation under scrutiny. Extensive experiments demonstrate that GS-IQA consistently outperforms existing MLLM-based IQA methods in fine-grained diagnosis, generalizes to external benchmarks without dataset-specific fine-tuning, and yields representations that transfer effectively to holistic quality prediction. Fine-grained diagnosis thus proves not a cost paid for interpretability, but a route to quality representations that serve both local diagnosis and image-level prediction.


\bibliography{iclr2027_conference}
\bibliographystyle{iclr2027_conference}

\end{document}